\documentclass[runningheads]{llncs}

\usepackage{accv}

\usepackage{accvabbrv}

\usepackage{graphicx}
\usepackage{booktabs}
\usepackage{multirow}
\usepackage{wrapfig}
\usepackage{algorithm}
\usepackage{algorithmic}

\usepackage[accsupp]{axessibility}  

\usepackage[pagebackref,breaklinks,colorlinks,citecolor=accvblue]{hyperref}

\usepackage{orcidlink}

\usepackage{colortbl}
\definecolor{tablehighlight}{HTML}{DAE6F2}
\newcommand{\scoredelta}[2]{$#1_{#2}$}
\newcommand{\bestscoredelta}[2]{$\mathbf{#1}_{#2}$}

\def\resultwooe{
\begin{table}[t]
\centering
\caption{OOD detection performance without Outlier Exposure (OE) using ResNet50. The final classification layer is used for the proposed method. RankWeight denotes the post-hoc RankWeight protocol. Throughout the tables, scores are reported as percentages, subscripts indicate differences from TWA in percentage points, and bold indicates the best result for each metric and experimental setting.}
\label{tb:result_wo_oe}
\footnotesize
\setlength{\tabcolsep}{2pt}
\resizebox{\textwidth}{!}{%
\begin{tabular}{llcccccc}
\toprule
\multirow{2}{*}{Protocol}
& \multirow{2}{*}{Method}
& \multicolumn{2}{c}{SVHN}
& \multicolumn{2}{c}{Textures}
& \multicolumn{2}{c}{Places} \\
\cmidrule(lr){3-4}\cmidrule(lr){5-6}\cmidrule(lr){7-8}
& & AUROC (\%)$\uparrow$ & FPR95 (\%)$\downarrow$
& AUROC (\%)$\uparrow$ & FPR95 (\%)$\downarrow$
& AUROC (\%)$\uparrow$ & FPR95 (\%)$\downarrow$ \\
\midrule
\multirow{3}{*}{MSP}
& FT
& 77.0 & 84.18
& 73.2 & 86.17
& 72.5 & 86.17 \\
& TWA
& 78.7 & 80.11
& 70.2 & 88.51
& 72.1 & 85.34 \\
& \cellcolor{tablehighlight}Ours
& \cellcolor{tablehighlight}\bestscoredelta{79.0}{+0.2} & \cellcolor{tablehighlight}\bestscoredelta{79.02}{-1.09}
& \cellcolor{tablehighlight}\bestscoredelta{74.3}{+4.0} & \cellcolor{tablehighlight}\bestscoredelta{85.59}{-2.92}
& \cellcolor{tablehighlight}\bestscoredelta{75.2}{+3.1} & \cellcolor{tablehighlight}\bestscoredelta{81.61}{-3.73} \\
\midrule
\multirow{3}{*}{Energy}
& FT
& 84.8 & 79.72
& \textbf{76.9} & \textbf{84.57}
& 75.0 & 84.49 \\
& TWA
& 83.5 & 79.80
& 71.2 & 89.41
& 74.2 & \textbf{80.75} \\
& \cellcolor{tablehighlight}Ours
& \cellcolor{tablehighlight}\bestscoredelta{88.4}{+4.9} & \cellcolor{tablehighlight}\bestscoredelta{68.28}{-11.52}
& \cellcolor{tablehighlight}\scoredelta{76.2}{+5.0} & \cellcolor{tablehighlight}\scoredelta{86.91}{-2.50}
& \cellcolor{tablehighlight}\bestscoredelta{76.1}{+1.9} & \cellcolor{tablehighlight}\scoredelta{80.78}{+0.03} \\
\midrule
\multirow{3}{*}{RankWeight}
& FT
& 77.8 & 84.32
& 65.3 & 92.87
& 71.6 & 87.61 \\
& TWA
& 74.7 & 88.47
& 63.9 & 92.98
& 72.8 & 83.42 \\
& \cellcolor{tablehighlight}Ours
& \cellcolor{tablehighlight}\bestscoredelta{88.3}{+13.6}
& \cellcolor{tablehighlight}\bestscoredelta{63.00}{-25.47}
& \cellcolor{tablehighlight}\bestscoredelta{73.0}{+9.1}
& \cellcolor{tablehighlight}\bestscoredelta{86.28}{-6.70}
& \cellcolor{tablehighlight}\bestscoredelta{73.3}{+0.5}
& \cellcolor{tablehighlight}\bestscoredelta{82.83}{-0.59} \\
\bottomrule
\end{tabular}
}
\end{table}
}

\def\resultwoe{
\begin{table}[t]
\centering
\caption{OOD detection performance with Outlier Exposure (OE) using ResNet50. The final classification layer is used for the proposed method. RankWeight denotes the post-hoc RankWeight protocol.}
\label{tb:result_w_oe}
\footnotesize
\setlength{\tabcolsep}{2pt}
\resizebox{\textwidth}{!}{%
\begin{tabular}{llcccccc}
\toprule
\multirow{2}{*}{Protocol}
& \multirow{2}{*}{Method}
& \multicolumn{2}{c}{SVHN}
& \multicolumn{2}{c}{Textures}
& \multicolumn{2}{c}{Places} \\
\cmidrule(lr){3-4}\cmidrule(lr){5-6}\cmidrule(lr){7-8}
& & AUROC (\%)$\uparrow$ & FPR95 (\%)$\downarrow$
& AUROC (\%)$\uparrow$ & FPR95 (\%)$\downarrow$
& AUROC (\%)$\uparrow$ & FPR95 (\%)$\downarrow$ \\
\midrule
\multirow{3}{*}{MSP}
& FT
& 79.4 & 80.22
& 74.4 & 83.99
& 75.3 & 82.10 \\
& TWA
& 80.5 & 77.27
& 72.1 & 86.91
& 71.1 & 85.08 \\
& \cellcolor{tablehighlight}Ours
& \cellcolor{tablehighlight}\bestscoredelta{82.9}{+2.5} & \cellcolor{tablehighlight}\bestscoredelta{70.34}{-6.93}
& \cellcolor{tablehighlight}\bestscoredelta{76.6}{+4.5} & \cellcolor{tablehighlight}\bestscoredelta{82.39}{-4.52}
& \cellcolor{tablehighlight}\bestscoredelta{77.3}{+6.2} & \cellcolor{tablehighlight}\bestscoredelta{78.62}{-6.46} \\
\midrule
\multirow{3}{*}{Energy}
& FT
& 84.2 & 75.80
& 77.4 & \textbf{81.33}
& 79.9 & 75.48 \\
& TWA
& 85.4 & 75.00
& 74.3 & 84.68
& 75.2 & 77.06 \\
& \cellcolor{tablehighlight}Ours
& \cellcolor{tablehighlight}\bestscoredelta{90.1}{+4.7} & \cellcolor{tablehighlight}\bestscoredelta{59.89}{-15.11}
& \cellcolor{tablehighlight}\bestscoredelta{79.2}{+4.9} & \cellcolor{tablehighlight}\scoredelta{81.49}{-3.19}
& \cellcolor{tablehighlight}\bestscoredelta{80.0}{+4.8} & \cellcolor{tablehighlight}\bestscoredelta{74.12}{-2.94} \\
\midrule
\multirow{3}{*}{RankWeight}
& FT
& 79.9 & 84.76
& 67.4 & 91.60
& 74.7 & 84.08 \\
& TWA
& 73.7 & 91.22
& 63.1 & 93.62
& 73.4 & 81.55 \\
& \cellcolor{tablehighlight}Ours
& \cellcolor{tablehighlight}\bestscoredelta{89.8}{+16.1}
& \cellcolor{tablehighlight}\bestscoredelta{56.44}{-34.78}
& \cellcolor{tablehighlight}\bestscoredelta{78.3}{+15.2}
& \cellcolor{tablehighlight}\bestscoredelta{80.43}{-13.19}
& \cellcolor{tablehighlight}\bestscoredelta{82.4}{+9.0}
& \cellcolor{tablehighlight}\bestscoredelta{71.24}{-10.31} \\
\bottomrule
\end{tabular}
}
\end{table}
}

\def\layerwise{
\begin{table}[t]
\centering
\caption{Target-layer analysis for dominant-component correction without OE on CIFAR-100/SVHN using ResNet50.}
\label{tb:layerwise}
\footnotesize
\setlength{\tabcolsep}{5pt}
\begin{tabular}{lcccc}
\toprule
\multirow{2}{*}{Method / Target}
& \multicolumn{2}{c}{MSP}
& \multicolumn{2}{c}{Energy} \\
\cmidrule(lr){2-3}\cmidrule(lr){4-5}
& AUROC (\%)$\uparrow$ & FPR95 (\%)$\downarrow$
& AUROC (\%)$\uparrow$ & FPR95 (\%)$\downarrow$ \\
\midrule
FT
& 77.0 & 84.18
& 84.8 & 79.72 \\
TWA
& 78.7 & 80.11
& 83.5 & 79.80 \\
\midrule
\multicolumn{5}{l}{\textit{Ours: target layer for dominant-component correction}} \\
\textit{all}
& \scoredelta{75.6}{-3.1} & \scoredelta{84.21}{+4.10}
& \scoredelta{84.2}{+0.7} & \scoredelta{78.49}{-1.31} \\
\textit{conv}
& \scoredelta{75.9}{-2.8} & \scoredelta{84.53}{+4.42}
& \scoredelta{80.5}{-3.0} & \scoredelta{86.14}{+6.34} \\
\textit{layer3}
& \scoredelta{76.7}{-2.0} & \scoredelta{85.56}{+5.45}
& \scoredelta{83.8}{+0.3} & \scoredelta{79.64}{-0.16} \\
\textit{layer4}
& \scoredelta{78.6}{-0.1} & \scoredelta{81.48}{+1.37}
& \scoredelta{86.9}{+3.4} & \scoredelta{71.56}{-8.24} \\
\cellcolor{tablehighlight}\textit{last}
& \cellcolor{tablehighlight}\bestscoredelta{79.0}{+0.2} & \cellcolor{tablehighlight}\bestscoredelta{79.02}{-1.09}
& \cellcolor{tablehighlight}\bestscoredelta{88.4}{+4.9} & \cellcolor{tablehighlight}\bestscoredelta{68.28}{-11.52} \\
\bottomrule
\end{tabular}
\end{table}
}

\def\averagingrange{
\begin{table}[t]
\centering
\caption{Effect of checkpoint averaging range on OOD detection without OE using ResNet50.}
\label{tb:avg_range}
\footnotesize
\setlength{\tabcolsep}{3pt}
\begin{tabular}{llcccc}
\toprule
\multirow{2}{*}{Range}
& \multirow{2}{*}{Method}
& \multicolumn{2}{c}{MSP}
& \multicolumn{2}{c}{Energy} \\
\cmidrule(lr){3-4}\cmidrule(lr){5-6}
& & AUROC (\%)$\uparrow$ & FPR95 (\%)$\downarrow$
& AUROC (\%)$\uparrow$ & FPR95 (\%)$\downarrow$ \\
\midrule
\multirow{2}{*}{(0,51)}
& TWA
& 71.1 & 85.60 & 79.4 & 89.30 \\
& \cellcolor{tablehighlight}Ours
& \cellcolor{tablehighlight}\bestscoredelta{74.3}{+3.2} & \cellcolor{tablehighlight}\bestscoredelta{83.18}{-2.42}
& \cellcolor{tablehighlight}\bestscoredelta{84.4}{+5.1} & \cellcolor{tablehighlight}\bestscoredelta{83.20}{-6.10} \\
\midrule
\multirow{2}{*}{(0,101)}
& TWA
& 74.1 & 83.50 & 83.5 & 79.80 \\
& \cellcolor{tablehighlight}Ours
& \cellcolor{tablehighlight}\bestscoredelta{78.3}{+4.2} & \cellcolor{tablehighlight}\bestscoredelta{81.28}{-2.22}
& \cellcolor{tablehighlight}\bestscoredelta{85.9}{+2.4} & \cellcolor{tablehighlight}\bestscoredelta{76.44}{-3.36} \\
\midrule
\multirow{2}{*}{(0,151)}
& TWA
& 70.8 & 84.83 & 79.9 & \textbf{83.93} \\
& \cellcolor{tablehighlight}Ours
& \cellcolor{tablehighlight}\bestscoredelta{75.8}{+5.0} & \cellcolor{tablehighlight}\bestscoredelta{83.03}{-1.80}
& \cellcolor{tablehighlight}\bestscoredelta{79.9}{+0.1} & \cellcolor{tablehighlight}\scoredelta{85.34}{+1.41} \\
\midrule
\multirow{2}{*}{(101,151)}
& TWA
& 70.1 & 87.66 & 79.4 & 86.24 \\
& \cellcolor{tablehighlight}Ours
& \cellcolor{tablehighlight}\bestscoredelta{75.8}{+5.6} & \cellcolor{tablehighlight}\bestscoredelta{83.45}{-4.21}
& \cellcolor{tablehighlight}\bestscoredelta{81.1}{+1.7} & \cellcolor{tablehighlight}\bestscoredelta{85.01}{-1.23} \\
\midrule
\multirow{2}{*}{(101,200)}
& TWA
& 70.6 & 87.23 & 78.9 & 86.29 \\
& \cellcolor{tablehighlight}Ours
& \cellcolor{tablehighlight}\bestscoredelta{75.6}{+5.1} & \cellcolor{tablehighlight}\bestscoredelta{83.81}{-3.42}
& \cellcolor{tablehighlight}\bestscoredelta{80.2}{+1.3} & \cellcolor{tablehighlight}\bestscoredelta{85.47}{-0.82} \\
\bottomrule
\end{tabular}
\end{table}
}

\def\resulttestacc{
\begin{wraptable}{r}{0.38\textwidth}
\vspace{-\intextsep}
\setlength{\abovecaptionskip}{0pt}
\centering
\caption{CIFAR-100 ID accuracy without OE.}
\label{tb:test_acc_lateral}
\footnotesize
\setlength{\tabcolsep}{5pt}
\begin{tabular}{lc}
\toprule
Method & Accuracy (\%)$\uparrow$ \\
\midrule
FT & 73.94 \\
TWA & 72.55 \\
\rowcolor{tablehighlight}
Ours & \bestscoredelta{75.04}{+2.49} \\
\bottomrule
\end{tabular}
\end{wraptable}
}

\def\resulttestaccoe{
\begin{wraptable}{r}{0.38\textwidth}
\vspace{-\intextsep}
\setlength{\abovecaptionskip}{0pt}
\centering
\caption{CIFAR-100 ID accuracy with OE.}
\label{tb:test_acc_oe}
\footnotesize
\setlength{\tabcolsep}{5pt}
\begin{tabular}{lc}
\toprule
Method & Accuracy (\%)$\uparrow$ \\
\midrule
FT & \textbf{73.94} \\
TWA & 69.94 \\
\rowcolor{tablehighlight}
Ours & \scoredelta{72.69}{+2.75} \\
\bottomrule
\end{tabular}
\end{wraptable}
}

\def\resultdatasetwooe{
\begin{table}[t]
\centering
\caption{Effect of backbone architectures on OOD detection without OE using CIFAR-100 as ID.}
\label{tb:dataset_centric_wo_oe}
\footnotesize
\setlength{\tabcolsep}{3pt}
\resizebox{\textwidth}{!}{%
\begin{tabular}{lllcccc}
\toprule
\multirow{2}{*}{Dataset}
& \multirow{2}{*}{Backbone}
& \multirow{2}{*}{Method}
& \multicolumn{2}{c}{MSP}
& \multicolumn{2}{c}{Energy} \\
\cmidrule(lr){4-5}\cmidrule(lr){6-7}
& & & AUROC (\%)$\uparrow$ & FPR95 (\%)$\downarrow$
& AUROC (\%)$\uparrow$ & FPR95 (\%)$\downarrow$ \\
\midrule
\multirow{9}{*}{SVHN}
& \multirow{3}{*}{VGG16BN}
& FT & \textbf{67.5} & 87.63 & 73.4 & 82.51 \\
& & TWA & 65.7 & 90.25 & 72.8 & 87.46 \\
& & \cellcolor{tablehighlight}Ours
& \cellcolor{tablehighlight}\scoredelta{66.8}{+1.1} & \cellcolor{tablehighlight}\bestscoredelta{84.88}{-5.37}
& \cellcolor{tablehighlight}\bestscoredelta{76.0}{+3.2} & \cellcolor{tablehighlight}\bestscoredelta{76.88}{-10.58} \\
\cmidrule(lr){2-7}
& \multirow{3}{*}{ResNet50}
& FT & 77.0 & 84.18 & 84.8 & 79.72 \\
& & TWA & 78.7 & 80.11 & 83.5 & 79.80 \\
& & \cellcolor{tablehighlight}Ours
& \cellcolor{tablehighlight}\bestscoredelta{79.0}{+0.2} & \cellcolor{tablehighlight}\bestscoredelta{79.02}{-1.09}
& \cellcolor{tablehighlight}\bestscoredelta{88.4}{+4.9} & \cellcolor{tablehighlight}\bestscoredelta{68.28}{-11.52} \\
\cmidrule(lr){2-7}
& \multirow{3}{*}{WideResNet40-2}
& FT & 69.5 & 89.14 & 79.9 & 81.50 \\
& & TWA & 74.6 & 86.63 & 79.7 & 86.27 \\
& & \cellcolor{tablehighlight}Ours
& \cellcolor{tablehighlight}\bestscoredelta{75.5}{+0.9} & \cellcolor{tablehighlight}\bestscoredelta{84.71}{-1.92}
& \cellcolor{tablehighlight}\bestscoredelta{83.9}{+4.3} & \cellcolor{tablehighlight}\bestscoredelta{81.23}{-5.04} \\
\midrule
\multirow{9}{*}{Textures}
& \multirow{3}{*}{VGG16BN}
& FT & \textbf{71.9} & 86.17 & 75.9 & 83.62 \\
& & TWA & 69.0 & 87.45 & 73.1 & 84.89 \\
& & \cellcolor{tablehighlight}Ours
& \cellcolor{tablehighlight}\scoredelta{70.2}{+1.3} & \cellcolor{tablehighlight}\bestscoredelta{84.20}{-3.25}
& \cellcolor{tablehighlight}\bestscoredelta{76.3}{+3.2} & \cellcolor{tablehighlight}\bestscoredelta{76.97}{-7.92} \\
\cmidrule(lr){2-7}
& \multirow{3}{*}{ResNet50}
& FT & 73.2 & 86.17 & \textbf{76.9} & \textbf{84.57} \\
& & TWA & 70.2 & 88.51 & 71.2 & 89.41 \\
& & \cellcolor{tablehighlight}Ours
& \cellcolor{tablehighlight}\bestscoredelta{74.3}{+4.0} & \cellcolor{tablehighlight}\bestscoredelta{85.59}{-2.92}
& \cellcolor{tablehighlight}\scoredelta{76.2}{+5.0} & \cellcolor{tablehighlight}\scoredelta{86.91}{-2.50} \\
\cmidrule(lr){2-7}
& \multirow{3}{*}{WideResNet40-2}
& FT & \textbf{72.8} & 86.33 & 72.5 & 89.47 \\
& & TWA & 72.1 & \textbf{86.17} & 75.2 & 84.89 \\
& & \cellcolor{tablehighlight}Ours
& \cellcolor{tablehighlight}\scoredelta{71.5}{-0.6} & \cellcolor{tablehighlight}\scoredelta{87.02}{+0.85}
& \cellcolor{tablehighlight}\bestscoredelta{76.9}{+1.7} & \cellcolor{tablehighlight}\bestscoredelta{81.33}{-3.56} \\
\midrule
\multirow{9}{*}{Places}
& \multirow{3}{*}{VGG16BN}
& FT & \textbf{72.4} & 85.76 & \textbf{73.8} & 84.38 \\
& & TWA & 69.2 & 88.92 & 70.1 & 86.37 \\
& & \cellcolor{tablehighlight}Ours
& \cellcolor{tablehighlight}\scoredelta{69.6}{+0.4} & \cellcolor{tablehighlight}\bestscoredelta{85.70}{-3.22}
& \cellcolor{tablehighlight}\scoredelta{70.4}{+0.3} & \cellcolor{tablehighlight}\bestscoredelta{83.30}{-3.07} \\
\cmidrule(lr){2-7}
& \multirow{3}{*}{ResNet50}
& FT & 72.5 & 86.17 & 75.0 & 84.49 \\
& & TWA & 72.1 & 85.34 & 74.2 & \textbf{80.75} \\
& & \cellcolor{tablehighlight}Ours
& \cellcolor{tablehighlight}\bestscoredelta{75.2}{+3.1} & \cellcolor{tablehighlight}\bestscoredelta{81.61}{-3.73}
& \cellcolor{tablehighlight}\bestscoredelta{76.1}{+1.9} & \cellcolor{tablehighlight}\scoredelta{80.78}{+0.03} \\
\cmidrule(lr){2-7}
& \multirow{3}{*}{WideResNet40-2}
& FT & 73.8 & 84.15 & 75.0 & 83.70 \\
& & TWA & \textbf{74.6} & \textbf{83.21} & 75.9 & 81.32 \\
& & \cellcolor{tablehighlight}Ours
& \cellcolor{tablehighlight}\scoredelta{73.1}{-1.5} & \cellcolor{tablehighlight}\scoredelta{84.11}{+0.90}
& \cellcolor{tablehighlight}\bestscoredelta{76.3}{+0.4} & \cellcolor{tablehighlight}\bestscoredelta{80.47}{-0.85} \\
\bottomrule
\end{tabular}
}
\end{table}
}

\def\resultmahalanobis{
\begin{table}[t]
\centering
\caption{Feature-based OOD detection using Mahalanobis scoring on ResNet50, with CIFAR-100 as ID. Features are extracted from the input to the final classification layer.}
\label{tb:result_mahalanobis}
\footnotesize
\setlength{\tabcolsep}{2pt}
\begin{tabular}{llcccccc}
\toprule
\multirow{2}{*}{OE} &
\multirow{2}{*}{Method} &
\multicolumn{2}{c}{SVHN} &
\multicolumn{2}{c}{Textures} &
\multicolumn{2}{c}{Places} \\
\cmidrule(lr){3-4} \cmidrule(lr){5-6} \cmidrule(lr){7-8}
& &
AUROC$\uparrow$ & FPR95$\downarrow$ &
AUROC$\uparrow$ & FPR95$\downarrow$ &
AUROC$\uparrow$ & FPR95$\downarrow$ \\
\midrule

\multirow{3}{*}{w/o OE}
& FT
& 71.2 & 77.20
& 86.6 & 43.94
& 51.5 & 95.95 \\

& TWA
& \textbf{84.1} & \textbf{55.20}
& 86.6 & 42.23
& \textbf{52.9} & \textbf{95.50} \\

& \cellcolor{tablehighlight}Ours
& \cellcolor{tablehighlight}\scoredelta{66.3}{-17.8}
& \cellcolor{tablehighlight}\scoredelta{86.29}{+31.09}
& \cellcolor{tablehighlight}\bestscoredelta{88.2}{+1.6}
& \cellcolor{tablehighlight}\bestscoredelta{41.22}{-1.01}
& \cellcolor{tablehighlight}\scoredelta{50.8}{-2.1}
& \cellcolor{tablehighlight}\scoredelta{95.56}{+0.06} \\

\midrule

\multirow{3}{*}{w/ OE}
& FT
& 71.2 & 79.69
& \textbf{88.6} & \textbf{38.30}
& 55.0 & 94.38 \\

& TWA
& \textbf{84.1} & \textbf{56.64}
& 85.6 & 42.71
& \textbf{58.9} & \textbf{93.66} \\

& \cellcolor{tablehighlight}Ours
& \cellcolor{tablehighlight}\scoredelta{69.4}{-14.7}
& \cellcolor{tablehighlight}\scoredelta{80.48}{+23.84}
& \cellcolor{tablehighlight}\scoredelta{86.4}{+0.8}
& \cellcolor{tablehighlight}\scoredelta{45.11}{+2.40}
& \cellcolor{tablehighlight}\scoredelta{51.6}{-7.3}
& \cellcolor{tablehighlight}\scoredelta{95.43}{+1.77} \\

\bottomrule
\end{tabular}%
\end{table}
}

\def\methodoverview{
\begin{figure}[t]
    \centering
    \includegraphics[width=\linewidth]{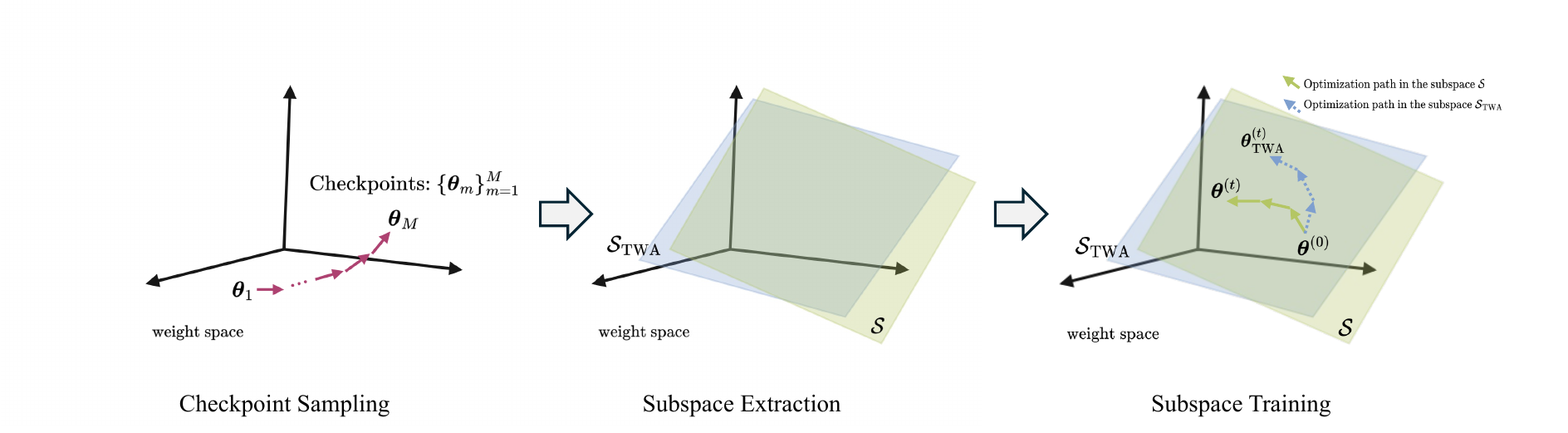}
    \caption{Overview of the proposed dominant-component-aware extension of TWA. For each checkpoint, leading rank-one components are extracted from selected weight matrices and added as correction directions to the TWA subspace. The final model is optimized in the expanded subspace using the ID classification objective.}
    \label{fig:method_overview}
\end{figure}
}

\def\losslandscape{
\begin{figure*}[t]
    \centering

    \begin{tabular}{ccc}
    FT &
    TWA &
    Ours \\

    \includegraphics[width=0.31\textwidth]{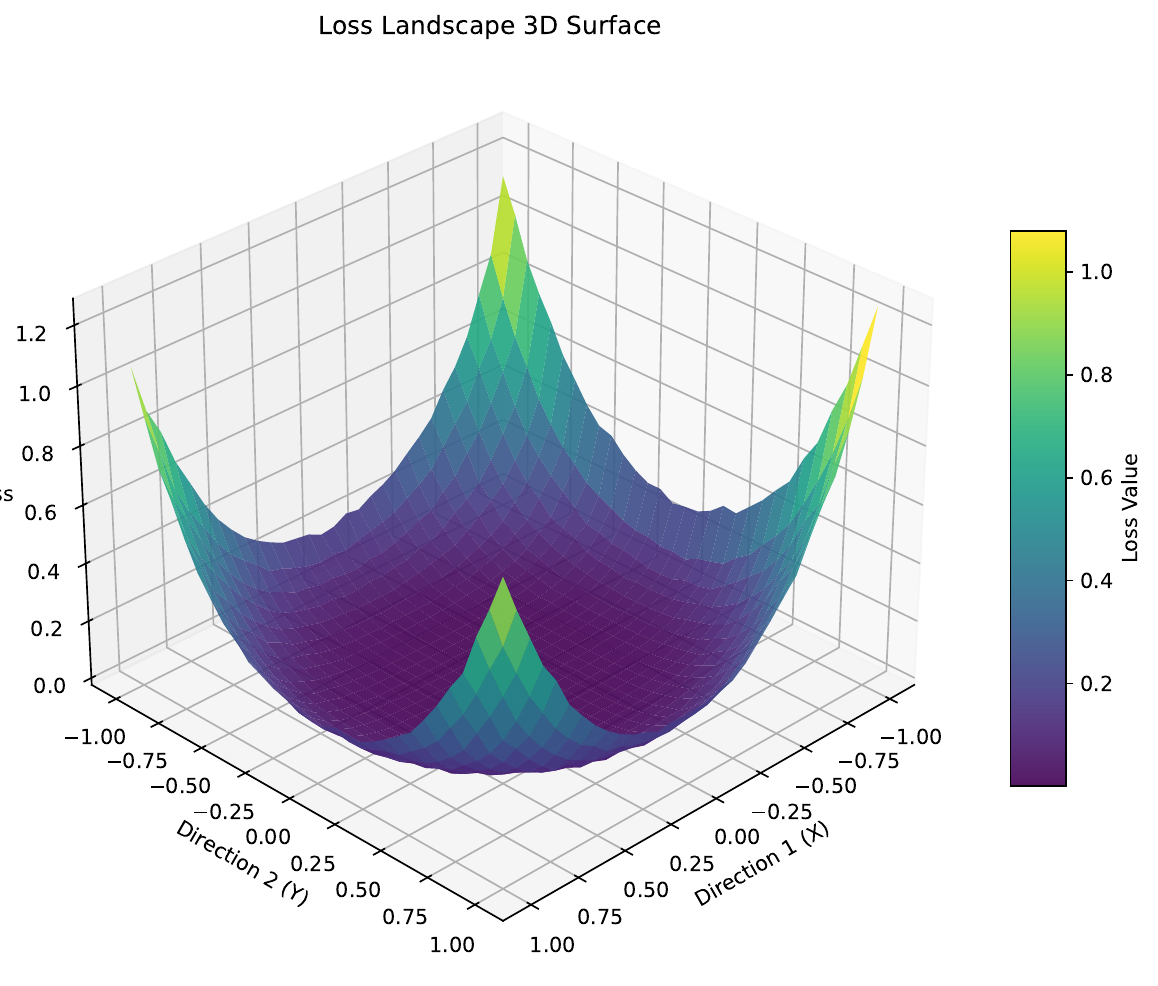} &
    \includegraphics[width=0.31\textwidth]{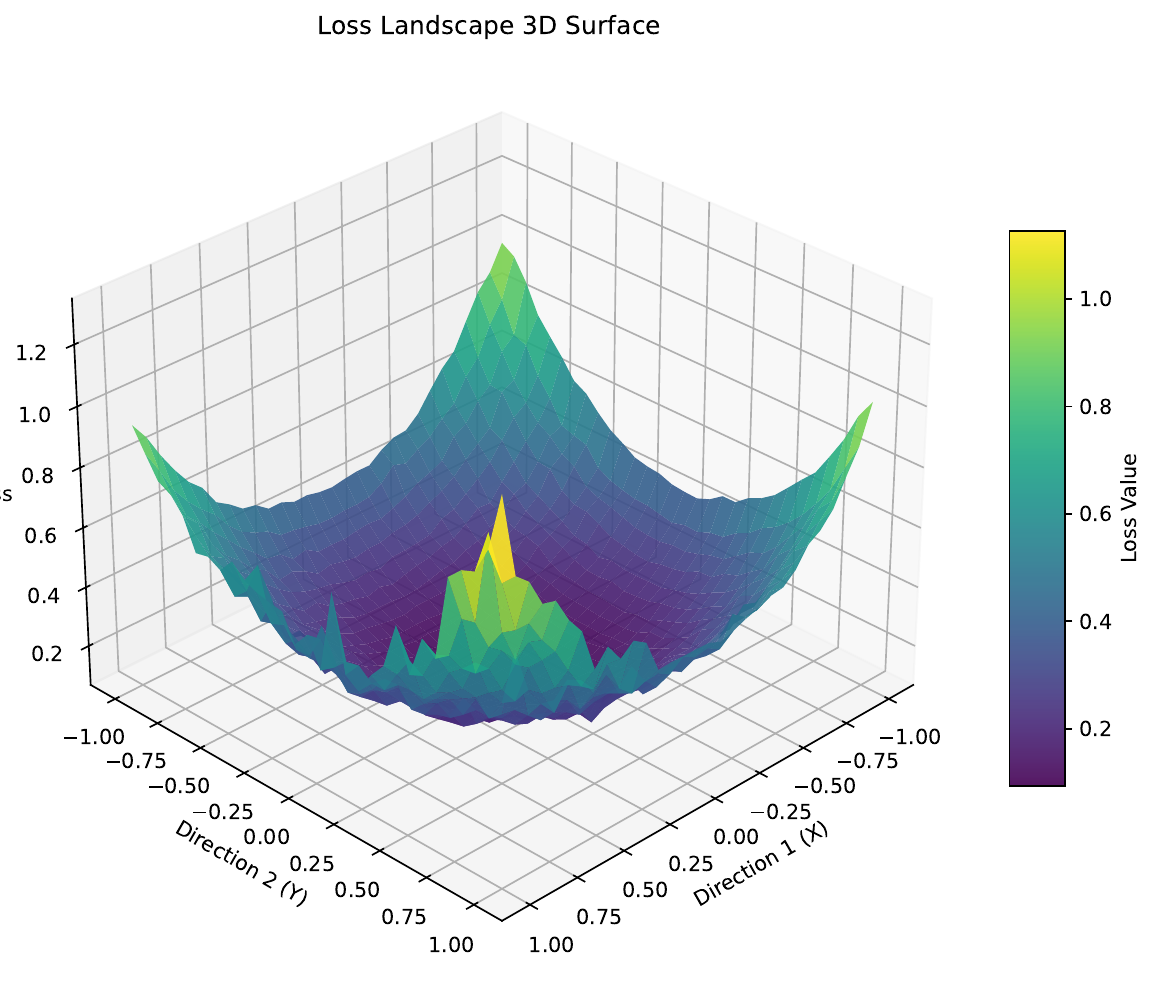} &
    \includegraphics[width=0.31\textwidth]{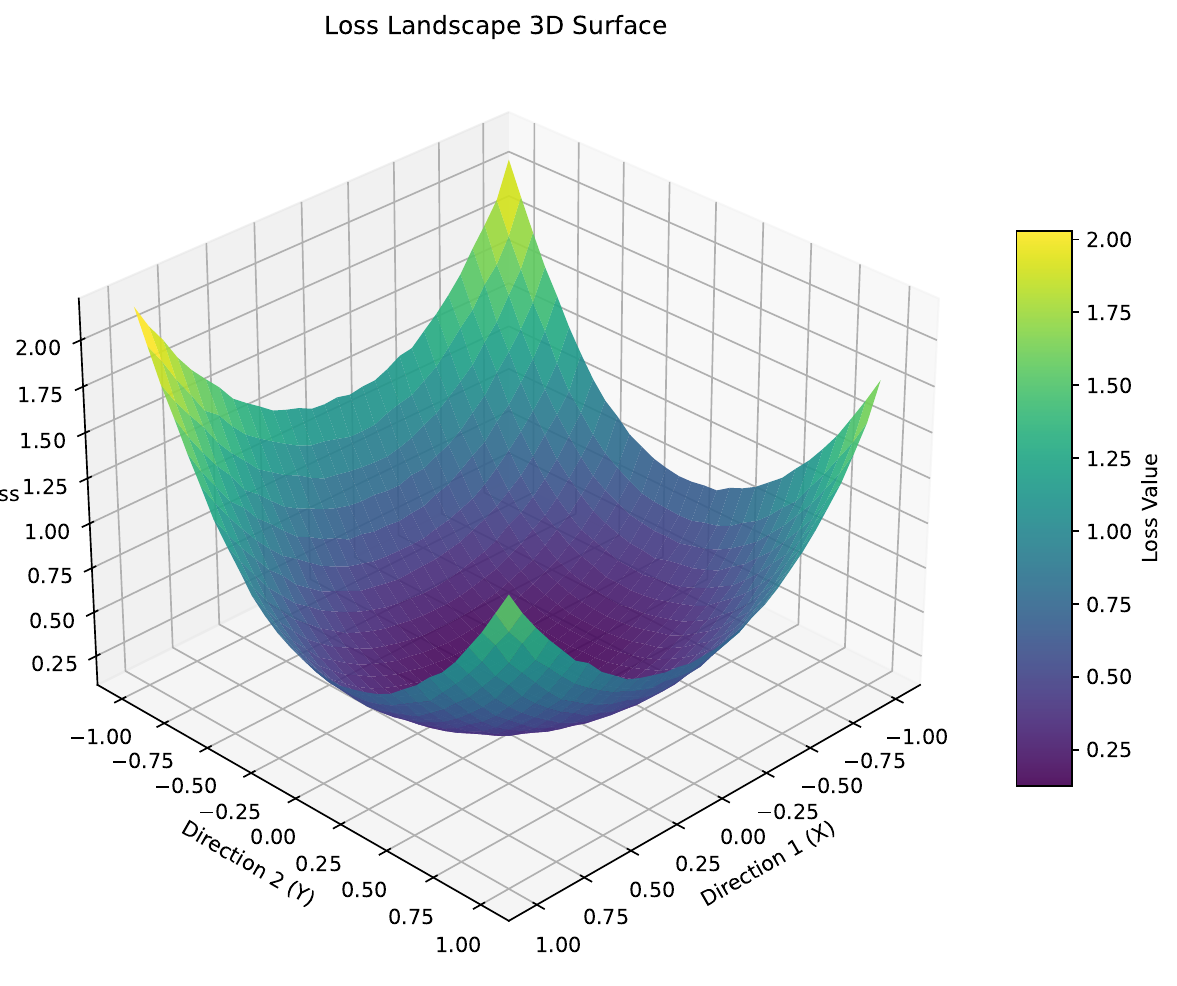}
    \end{tabular}

    \caption{Loss landscape visualization without Outlier Exposure (OE). The proposed method expands the TWA search subspace while retaining the flat-solution behavior of weight averaging.}
    \label{fig:loss_landscape}
\end{figure*}
}

\begin{document}

\title{Rank-1 Weight-Corrected Trainable Weight Averaging for Out-of-Distribution Detection} 

\titlerunning{Rank-1 Weight-Corrected Trainable Weight Averaging for OOD Detection}

\author{Akito Shinohara \and Hiroaki Aizawa}

\authorrunning{A. Shinohara and H. Aizawa}

\institute{
Graduate School of Advanced Science and Engineering, \\
Hiroshima University, Japan
\\
\email{\{m252486,hiroaki-aizawa\}@hiroshima-u.ac.jp}
}

\maketitle

\begin{abstract}
Out-of-distribution (OOD) detection identifies inputs outside the training distribution. Rank-based analyses such as RankFeat and RankWeight suggest that OOD overconfidence is closely related to dominant low-rank components in feature and weight spaces. Since weight averaging operates directly in parameter space and improves model generalization, we consider it for OOD detection from a dominant-component perspective. Trainable Weight Averaging (TWA) assigns a single coefficient to each checkpoint, limiting independent control of dominant weight components. We extend TWA with leading rank-one weight components extracted by Singular Value Decomposition as correction directions, whose contributions are learned through the in-distribution classification objective. On ResNet50, our method improves over fine-tuning and standard TWA across the evaluated datasets with MSP and RankWeight, both with and without Outlier Exposure. Mahalanobis-based evaluation yields mixed results, indicating that the benefit depends on the OOD scoring protocol. Target-layer analyses with MSP and Energy favor correction near the classifier.

\keywords{Out-of-Distribution Detection \and Trainable Weight Averaging \and Subspace Optimization}
\end{abstract}

\section{Introduction}
\label{sec:intro}
Reliable prediction requires models to maintain discriminative performance on known classes while identifying unknown data at test time. In real-world environments, deep neural networks often encounter inputs that are not represented in the training distribution. Such inputs can induce overconfident predictions even when they do not belong to any known class, which is problematic in safety-critical applications. Out-of-distribution (OOD) detection \cite{ood} addresses this problem by determining whether an input lies outside the training distribution.

A broad range of OOD detection methods has been developed using model outputs or intermediate representations. Post-hoc scoring methods, such as Maximum Softmax Probability (MSP) \cite{msp} and Energy Score \cite{energy}, use confidence or logit-derived scores to separate in-distribution (ID) and OOD samples. Feature-space methods, such as Mahalanobis Distance \cite{mahalanobis} and ReAct \cite{react}, exploit statistical properties of learned representations. In addition to these post-hoc scoring and feature-level perspectives, recent rank-based analyses such as RankFeat and RankWeight \cite{rankfeat} suggest that dominant low-rank components in feature and weight spaces are closely related to OOD behavior and model performance. These findings indicate that OOD overconfidence is not only a property of output scores, but is also linked to the geometry of representations and model parameters.

This parameter-space perspective naturally connects to weight averaging methods. Weight averaging has been widely studied as a way to improve generalization by operating directly on model parameters, including Stochastic Weight Averaging (SWA) \cite{swa}, Model Soup \cite{model_soup}, SWAD \cite{swad}, and Trainable Weight Averaging (TWA) \cite{twa}. Among these methods, TWA learns averaging coefficients in a subspace spanned by candidate weights, providing a natural parameter-space mechanism for optimizing combinations of checkpoints. However, standard TWA assigns a single coefficient to each checkpoint. As a result, dominant low-rank weight components and the remaining checkpoint parameters are controlled together, making it difficult to adjust components that may contribute to OOD overconfidence without also changing components useful for ID classification.

Motivated by this limitation, we propose a dominant-component-aware extension of TWA. For each checkpoint, we apply Singular Value Decomposition (SVD) to selected weight matrices, extract the leading rank-one component, and introduce the resulting correction direction into the TWA subspace. The correction magnitude is learned with the ID classification objective, enabling adaptive control of dominant weight components without hard removal, architectural modification, or inference-time overhead. Since the proposed method operates in weight space, it can also be combined with data-driven OOD regularization such as Outlier Exposure (OE)\cite{oe}. We evaluate the proposed method on OOD detection tasks and compare it with conventional fine-tuning and standard TWA. We further analyze checkpoint ranges, target layers for the correction, compatibility with OE, and backbone architectures to clarify when dominant-component control is effective.

The main contributions of this work are summarized as follows:

\begin{itemize}
\item We propose a dominant-component-aware extension of TWA that augments the trainable weight subspace with correction directions derived from leading rank-one weight components.

\item We enable independent control of dominant weight components and residual checkpoint components while retaining the standard inference cost and network architecture.

\item We demonstrate improvements over fine-tuning and standard TWA with MSP and RankWeight on ResNet50, both with and without OE, while identifying mixed results with Mahalanobis scoring. Target-layer analyses with MSP and Energy show the strongest performance when correction is applied to the final classification layer.

\end{itemize}

\section{Related Work}
\label{sec:related_work}
OOD detection has been studied from several complementary perspectives. Post-hoc scoring methods, such as Maximum Softmax Probability (MSP) \cite{msp} and Energy Score \cite{energy}, detect OOD samples using confidence or logit-derived scores without changing the trained model. Feature-space methods, such as Mahalanobis Distance \cite{mahalanobis} and ReAct \cite{react}, exploit statistics or activations of intermediate representations. Other methods improve OOD detection during training, for example by regularizing features \cite{vos} or using auxiliary outlier data through Outlier Exposure \cite{oe}. These approaches mainly act on outputs, representations, or data supervision, whereas our work focuses on the structure of model parameters induced by weight averaging.

Weight averaging is a parameter-space approach for improving generalization and robustness. Stochastic Weight Averaging (SWA) \cite{swa} averages checkpoints collected during training, Model Soups \cite{model_soup} merge independently trained models, SWAD \cite{swad} searches for flatter solutions, and Trainable Weight Averaging (TWA) \cite{twa} learns averaging coefficients in a subspace spanned by candidate weights. Separately, rank-based analyses have shown that dominant spectral components are closely related to OOD behavior: RankFeat \cite{rankfeat} reports dominant rank-one components in OOD feature representations, and RankWeight \cite{rankfeat} shows that controlling dominant low-rank components in weight matrices can improve model performance. However, standard TWA controls each checkpoint with a single coefficient and does not separately adjust such dominant weight components. Our method addresses this gap by augmenting the TWA subspace with correction directions derived from leading rank-one weight components.

\section{Proposed Method}
\label{sec:method}
This section presents our dominant-component-aware extension of TWA, whose goal is to improve OOD detection by adaptively controlling dominant weight components. The key idea is to augment the TWA search subspace with correction directions extracted from leading rank-one checkpoint components, as summarized in Section~\ref{subsec:overview}. We first motivate this design in Section~\ref{subsec:problem_motivation}, then describe the formulation, component extraction, subspace optimization, and adaptive-control interpretation in Sections~\ref{subsec:overview}, \ref{subsec:dc_extract}, \ref{subsec:coefficient_optimization}, and \ref{subsec:adaptive_suppression}, respectively.

\methodoverview

\subsection{Motivation}
\label{subsec:problem_motivation}
RankWeight reports that the rank-one component associated with the largest singular value of a weight matrix can contribute more strongly to overconfident OOD predictions than to ID classification performance~\cite{rankfeat}. A direct way to use this observation is to remove the dominant component from the weight matrix. However, such hard removal may also discard information that is useful for classifying ID samples.

We therefore do not remove dominant weight components deterministically. Instead, we introduce them into the Trainable Weight Averaging (TWA) search subspace as correction directions and learn their contributions with the ID classification objective. This allows the model to adaptively control dominant weight components while preserving discriminative information needed for ID classification. The proposed method does not modify the network architecture and incurs no additional inference-time cost because the final model is still represented by a single set of parameters.

\subsection{Overview}
\label{subsec:overview}
We consider a $K$-class ID classification problem. Let the ID training dataset be $\mathcal{D}_{\mathrm{in}}=\{(\mathbf{x}_i,y_i)\}_{i=1}^{N}$, and let $f(\cdot;\boldsymbol{\theta})$ be a neural network with parameters $\boldsymbol{\theta} \in \mathbb{R}^{D}$. At test time, an input may come either from the ID distribution $P_{\mathrm{in}}$ or from an unknown OOD distribution $P_{\mathrm{out}}$, whose samples do not belong to the known classes. For OOD detection, the trained model assigns a score $S(\mathbf{x};\boldsymbol{\theta})$ to an input $\mathbf{x}$, where a higher score indicates a higher likelihood of being OOD. Given a threshold $\tau$, the decision function is
\begin{align}
    g(\mathbf{x};\boldsymbol{\theta},\tau)
    =
    \mathbb{I}
    \left[
    S(\mathbf{x};\boldsymbol{\theta})\geq\tau
    \right],
\end{align}
where $\mathbb{I}[\cdot]$ is the indicator function.
Thus, $g=1$ when the OOD score of $\mathbf{x}$ is at least $\tau$, and $g=0$ otherwise; the threshold $\tau$ determines the ID/OOD decision boundary.

The proposed method modifies how the model parameters $\boldsymbol{\theta}$ are obtained before applying such a detector.
Following TWA, we learn the parameters in a checkpoint subspace using the ID classification objective.
Let $\ell$ denote the classification loss, and define the empirical objective as
\begin{align}
    \mathcal{L}(\boldsymbol{\theta})
    =
    \frac{1}{N}
    \sum_{i=1}^{N}
    \ell
    \left(
    f(\mathbf{x}_i;\boldsymbol{\theta}),
    y_i
    \right).
\end{align}

Let $\boldsymbol{\theta}_1,\boldsymbol{\theta}_2,\ldots,\boldsymbol{\theta}_M$ denote $M$ checkpoints saved during training, where $\boldsymbol{\theta}_m \in \mathbb{R}^{D}$ is the vectorized parameter of checkpoint $m$. Standard TWA optimizes model parameters in the subspace spanned by these checkpoints:
\begin{align}
    \mathcal{S}_{\mathrm{TWA}}
    =
    \mathrm{span}
    \left\{
    \boldsymbol{\theta}_1,\ldots,\boldsymbol{\theta}_M
    \right\}.
\end{align}
In contrast, the proposed method extracts dominant weight components from selected weight matrices of each checkpoint and constructs a dominant-component vector $\mathbf{r}_m \in \mathbb{R}^{D}$. We then expand the TWA search subspace using the checkpoints and the correction directions $-\mathbf{r}_m$:
\begin{align}
    \mathcal{S}
    =
    \mathrm{span}
    \left\{
    \boldsymbol{\theta}_1,\ldots,\boldsymbol{\theta}_M,
    -\mathbf{r}_1,\ldots,-\mathbf{r}_M
    \right\}.
\end{align}

\emph{Subspace Extraction} in Figure~\ref{fig:method_overview} illustrates the subspace expansion. Since $\mathcal{S}_{\mathrm{TWA}}\subseteq\mathcal{S}$, the proposed method can be interpreted as an extension of TWA that augments the search space with directions corresponding to dominant weight components. The following subsections describe how the dominant components are extracted and how the expanded-subspace coefficients are optimized.

\subsection{Dominant Weight Component Extraction}
\label{subsec:dc_extract}
Following the checkpoint-based pipeline of TWA, we process each saved checkpoint and extract dominant weight components from selected target layers. For notational simplicity, we describe the extraction for one selected weight matrix in checkpoint $m$ and denote it by $\mathbf{W}_m \in \mathbb{R}^{d'\times d}$, where $d'$ is the output-channel or output-unit dimension and $d$ is the remaining input dimension. The same operation is applied to all selected target layers. For a fully connected layer, the weight matrix is used directly. For a convolutional layer, the weight tensor is reshaped into a two-dimensional matrix by preserving the output-channel dimension and flattening the remaining dimensions. Singular Value Decomposition (SVD) decomposes $\mathbf{W}_m$ as
\begin{align}
    \mathbf{W}_m
    =
    \mathbf{U}_m\boldsymbol{\Sigma}_m \mathbf{V}_m^{\top},
\end{align}
where $\mathbf{U}_m\in\mathbb{R}^{d'\times d'}$ and $\mathbf{V}_m\in\mathbb{R}^{d\times d}$ are the left and right singular-vector matrices, respectively, and $\boldsymbol{\Sigma}_m\in\mathbb{R}^{d'\times d}$ is the singular-value matrix. The diagonal entries of $\boldsymbol{\Sigma}_m$ are the singular values satisfying $\sigma_{m,1} \geq \sigma_{m,2} \geq \cdots$. The columns of $\mathbf{U}_m$ and $\mathbf{V}_m$ are the left and right singular vectors, respectively. We define the rank-one component corresponding to the largest singular value as the dominant weight component:
\begin{align}
    \mathbf{C}_m
    =
    \sigma_{m,1}\mathbf{u}_{m,1}\mathbf{v}_{m,1}^{\top}.
\end{align}
For convolutional layers, the extracted component is reshaped back to the original tensor shape. The dominant-component vector $\mathbf{r}_m \in \mathbb{R}^{D}$ is constructed by placing the extracted components at their corresponding parameter locations and setting all remaining dimensions to zero. Thus, $\mathbf{r}_m$ has the same dimensionality as $\boldsymbol{\theta}_m$ and stores the dominant weight components extracted from checkpoint $m$. In the following subspace optimization, we add $-\mathbf{r}_m$ to the TWA subspace as a correction direction.

\subsection{Subspace Optimization}
\label{subsec:coefficient_optimization}
Following TWA, we construct an orthonormal basis $\mathbf{P}=[\mathbf{e}_1,\ldots,\mathbf{e}_{d_{\mathcal S}}]$ of the expanded subspace $\mathcal{S}$, where $d_{\mathcal S}=\dim(\mathcal{S})$. A parameter vector in the subspace is represented as $\boldsymbol{\theta}(\boldsymbol{\gamma})=\mathbf{P}\boldsymbol{\gamma}$ using low-dimensional coordinates $\boldsymbol{\gamma}\in\mathbb{R}^{d_{\mathcal S}}$.

\emph{Subspace Training} in Figure~\ref{fig:method_overview} illustrates optimization in the expanded subspace. As in TWA, the coefficient optimization is implemented as subspace training: gradients are first computed in the full parameter space and then projected onto the search subspace. The coefficients are learned from ID training data by solving
\begin{align}
    \boldsymbol{\gamma}^{*}
    =
    \mathop{\rm argmin}_{\boldsymbol{\gamma}}
    \frac{1}{N}
    \sum_{i=1}^{N}
    \ell
    \left(
    f
    \left(
    \mathbf{x}_i;
    \mathbf{P}\boldsymbol{\gamma}
    \right),
    y_i
    \right)
    +
    \frac{\lambda}{2}
    \left\|
    \boldsymbol{\gamma}
    \right\|_2^2,
\end{align}
where $\lambda$ is the regularization strength for the coefficients. In practice, rather than explicitly updating $\boldsymbol{\gamma}$, we project the gradient with respect to the current parameter $\boldsymbol{\theta}^{(t)}$ onto the subspace spanned by $\mathbf{P}$. With learning rate $\eta$, the update is written as
\begin{align}
    \boldsymbol{\theta}^{(t+1)}
    =
    (1-\eta\lambda)\boldsymbol{\theta}^{(t)}
    -
    \eta \mathbf{P}\mathbf{P}^{\top}
    \nabla_{\boldsymbol{\theta}}
    \mathcal{L}
    \left(
    \boldsymbol{\theta}^{(t)}
    \right).
    \label{eq:projected_update}
\end{align}
The final model parameter is denoted by $\boldsymbol{\theta}^{*}$. Since $\mathbf{P}$ is a basis of $\mathcal{S}$, there exist coefficients $\alpha_m,\beta_m\in\mathbb{R}$ such that the final parameter can be equivalently expressed as
\begin{align}
    \boldsymbol{\theta}^{*}
    =
    \sum_{m=1}^{M}
    \alpha_m\boldsymbol{\theta}_m
    -
    \sum_{m=1}^{M}
    \beta_m \mathbf{r}_m.
    \label{eq:dcs_weight}
\end{align}
Here, $\boldsymbol{\gamma}$ denotes coordinates in the orthonormal basis, while $\alpha_m$ and $\beta_m$ denote coefficients in an equivalent representation using the original checkpoints and dominant-component vectors.

\subsection{Adaptive Control of Dominant Weight Components}
\label{subsec:adaptive_suppression}
To interpret the effect of the proposed subspace, we decompose each checkpoint as $\boldsymbol{\theta}_m=\mathbf{q}_m+\mathbf{r}_m$, where $\mathbf{q}_m=\boldsymbol{\theta}_m-\mathbf{r}_m$ denotes the residual parameter after subtracting the dominant weight component in the selected layers. Equation~\eqref{eq:dcs_weight} can then be rewritten as
\begin{equation}
\begin{aligned}
    \boldsymbol{\theta}^{*}
    &=
    \sum_{m=1}^{M}
    \alpha_m
    \left(
    \mathbf{q}_m+\mathbf{r}_m
    \right)
    -
    \sum_{m=1}^{M}
    \beta_m \mathbf{r}_m \\
    &=
    \sum_{m=1}^{M}
    \alpha_m \mathbf{q}_m
    +
    \sum_{m=1}^{M}
    \left(
    \alpha_m-\beta_m
    \right)
    \mathbf{r}_m.
\end{aligned}
\label{eq:dcs_decomposition}
\end{equation}
In standard TWA, the residual component $\mathbf{q}_m$ and the dominant weight component $\mathbf{r}_m$ are controlled by the same coefficient. In contrast, the proposed method gives the dominant component an effective coefficient of $\alpha_m-\beta_m$, allowing its contribution to be adjusted independently from the residual component. Thus, the expanded subspace can search for solutions that retain components useful for ID classification while reducing the contribution of dominant weight components when such reduction is beneficial. The negative sign in $-\mathbf{r}_m$ should be interpreted as introducing a correction direction, not as deterministic component removal. Since $\mathrm{span}\{\mathbf{r}_m\}=\mathrm{span}\{-\mathbf{r}_m\}$, suppression is determined by the learned coefficients in the expanded TWA subspace.

Algorithm~\ref{alg:rank1_twa} summarizes the overall procedure. The proposed method follows the checkpoint-subspace optimization pipeline of TWA, but expands the search subspace with correction directions extracted from leading rank-one weight components.

\begin{algorithm}[t]
\caption{Rank-1 Weight-Corrected TWA}
\label{alg:rank1_twa}
\small
\begin{algorithmic}[1]
\REQUIRE ID training set $\mathcal{D}_{\mathrm{in}}$, number of checkpoints $M$, target layer set $\mathcal{T}$, loss $\ell$, learning rate $\eta$, regularization coefficient $\lambda$
\ENSURE Final parameter $\boldsymbol{\theta}^{*}$ for OOD scoring
\STATE \textbf{\# Checkpoint Sampling}
\STATE Collect $M$ checkpoints $\{\boldsymbol{\theta}_m\}_{m=1}^{M}$ during training.
\STATE \textbf{\# Subspace Extraction}
\FOR{$m=1,\ldots,M$}
    \STATE Initialize $\mathbf{r}_m\leftarrow\mathbf{0}\in\mathbb{R}^{D}$.
    \FOR{each target layer $l\in\mathcal{T}$}
        \STATE Extract rank-one component: $\mathbf{C}_{m}^{(l)}\leftarrow\sigma_{m,1}^{(l)}\mathbf{u}_{m,1}^{(l)}\mathbf{v}_{m,1}^{(l)\top}$.
        \STATE Place $\mathbf{C}_{m}^{(l)}$ in the corresponding entries of $\mathbf{r}_m$.
    \ENDFOR
\ENDFOR
\STATE Construct $\mathcal{S}=\mathrm{span}\{\boldsymbol{\theta}_1,\ldots,\boldsymbol{\theta}_M,-\mathbf{r}_1,\ldots,-\mathbf{r}_M\}$.
\STATE Compute an orthonormal basis $\mathbf{P}$ of $\mathcal{S}$.
\STATE \textbf{\# Subspace Training}
\STATE Initialize $\boldsymbol{\theta}^{(0)}$ in $\mathcal{S}$ and set $t\leftarrow 0$.
\WHILE{not converged}
    \STATE Sample a mini-batch $\mathcal{B}\subset\mathcal{D}_{\mathrm{in}}$.
    \STATE Compute $\mathbf{g}^{(t)}\leftarrow\nabla_{\boldsymbol{\theta}}\mathcal{L}_{\mathcal{B}}(\boldsymbol{\theta}^{(t)})$.
    \STATE Project the gradient: $\bar{\mathbf{g}}^{(t)}\leftarrow\mathbf{P}\mathbf{P}^{\top}\mathbf{g}^{(t)}$.
    \STATE Update $\boldsymbol{\theta}^{(t+1)}\leftarrow(1-\eta\lambda)\boldsymbol{\theta}^{(t)}-\eta\bar{\mathbf{g}}^{(t)}$.
    \STATE Set $t\leftarrow t+1$.
\ENDWHILE
\STATE Set $\boldsymbol{\theta}^{*}\leftarrow\boldsymbol{\theta}^{(t)}$.
\RETURN $\boldsymbol{\theta}^{*}$
\end{algorithmic}
\end{algorithm}

\section{Experiments}
\label{sec:exp}

We evaluate the proposed rank-1 weight-corrected TWA on image classification-based OOD detection tasks. The experiments are designed to examine whether the proposed correction improves OOD detection over fine-tuning and standard TWA, whether it remains effective when combined with Outlier Exposure, and how its behavior changes with the averaging range, target layers, and backbone architecture.

\subsection{Experimental Setup}
\label{subsec:setup}

\paragraph{Datasets.}
We use CIFAR-100\cite{cifar100} as the in-distribution (ID) dataset and evaluate OOD detection on three test datasets with different visual characteristics: SVHN\cite{svhn}, Textures (DTD)\cite{dtd}, and Places (Places365)\cite{places365}. In the Outlier Exposure (OE) setting~\cite{oe}, TinyImages300K\cite{tinyimages300k} is used as an auxiliary outlier dataset during training. The auxiliary outlier data are separate from the OOD test datasets.

\paragraph{Training Details.}
ResNet50\cite{resnet} is used as the primary backbone. We also evaluate VGG16BN\cite{vgg16,bn} and WideResNet-40-2\cite{wideresnet} to examine whether the proposed correction is robust across architectures. Each backbone is trained for 200 epochs using SGD with a batch size of 128, an initial learning rate of 0.1, a momentum of 0.9, and a weight decay of $1\times10^{-4}$. The learning rate is decayed by a factor of 10 at epochs 100 and 150 using a multi-step learning rate scheduler. Checkpoints are saved at the end of every epoch. Following TWA~\cite{twa}, we regard the saved checkpoints as candidate weights. In the comparison stage, FT, TWA, and the proposed method are optimized for an additional 50 epochs; TWA and the proposed method construct their search subspaces from the selected candidate weights.

\paragraph{Proposed Variants.}
For the proposed method, Singular Value Decomposition (SVD) is applied to the selected weight matrices of each candidate checkpoint, and the leading rank-one component is used to construct the correction direction. We investigate five choices of target layers for dominant-component extraction: all layers (\textit{all}), convolutional layers only (\textit{conv}), \textit{layer3}, \textit{layer4}, and the final classification layer (\textit{last}). Unless otherwise stated, we report the variant using the final classification layer.

\paragraph{Evaluation.}
We evaluate OOD detection using Maximum Softmax Probability (MSP)~\cite{msp} and Energy Score~\cite{energy}. We also include RankWeight~\cite{rankfeat}, a post-hoc baseline that removes dominant weight components after training. To assess feature-based detection, we additionally evaluate Mahalanobis distance~\cite{mahalanobis} using features extracted from the input to the final classification layer on ResNet50, with and without OE. Performance is measured by AUROC and FPR95, where higher AUROC and lower FPR95 indicate better OOD detection. We compare the proposed method with conventional fine-tuning (FT) and Trainable Weight Averaging (TWA), and conduct ablations over the checkpoint range, target layers, and backbone architecture.

\resultwooe
\subsection{Main Results}
\label{subsec:main_result}
\paragraph{OOD Detection without Auxiliary Outlier Data.}
We first evaluate the proposed method in the standard setting where no auxiliary outlier data are used during training. This setting directly tests whether expanding the TWA subspace with rank-1 weight-component correction can improve OOD detection from ID training data alone.

Table~\ref{tb:result_wo_oe} reports the OOD detection results without OE using ResNet50 as the backbone, organized by OOD scoring protocol. Across MSP and RankWeight, the proposed method consistently improves TWA on all OOD datasets in both AUROC and FPR95. For Energy Score, it improves AUROC on all datasets and substantially reduces FPR95 on SVHN (79.80\% to 68.28\%) and Textures (89.41\% to 86.91\%), while the FPR95 on Places is essentially unchanged from TWA (80.75\% vs. 80.78\%). Compared with FT, the proposed method is strongest in the MSP and RankWeight protocols, whereas FT remains competitive for Energy-based detection on Textures. These results indicate that the rank-1 correction improves the TWA solution most consistently for confidence- and rank-based OOD detection without auxiliary outlier data.

\paragraph{ID Classification Performance.}
\resulttestacc
Since the proposed correction is optimized with the ID classification objective, we also examine whether the OOD improvements come at the cost of ID accuracy. Table~\ref{tb:test_acc_lateral} reports CIFAR-100 ID classification accuracy in the same without-OE setting as Table~\ref{tb:result_wo_oe}. Standard TWA decreases accuracy compared with FT (72.55\% vs. 73.94\%), whereas the proposed method improves it to 75.04\%, corresponding to a 2.49-point gain over TWA. These results indicate that the proposed rank-1 correction improves OOD detection while preserving, and in this setting improving, ID classification performance.

\subsection{Analysis of the Proposed Method}
\label{subsec:analysis}

\paragraph{Effect of Averaging Range.}
Since the checkpoint averaging range determines the candidate weights that span the TWA subspace, we examine whether the proposed correction depends strongly on a particular range. Table~\ref{tb:avg_range} reports the results without OE using different checkpoint ranges. The proposed method improves MSP-based AUROC and FPR95 over TWA for all ranges, and also improves Energy-based AUROC across all ranges. Energy-based FPR95 is improved in most ranges, with the exception of the range (0,151). The strongest overall performance is obtained from (0,101), while the late-only ranges (101,151) and (101,200) are not consistently superior. These results suggest that the benefit of rank-1 correction is not simply due to using later checkpoints, but comes from improving the TWA subspace across a range of checkpoint selections.

\averagingrange

\paragraph{Target Layers for Dominant-Component Correction.}
We next analyze how the choice of target layers affects the proposed correction. Table~\ref{tb:layerwise} reports results without OE when the leading rank-one components are extracted from different parts of ResNet50. Applying the correction to the final classification layer achieves the best performance for both MSP and Energy Score, while applying it to layer4 is also effective, especially for Energy-based detection. In contrast, applying the correction to all layers or only convolutional layers gives limited gains and can even degrade MSP-based detection. These results indicate that rank-1 weight-component correction is most effective when localized near the classifier, rather than uniformly applied across the whole feature extractor.

\layerwise

\paragraph{Effect of Backbone Architectures.}
We further examine whether the proposed correction remains effective across different backbone architectures. Table~\ref{tb:dataset_centric_wo_oe} reports the results without OE for VGG16BN, ResNet50, and WideResNet-40-2. On ResNet50, the proposed method gives the most consistent improvements over TWA across datasets and scoring protocols, especially for Energy-based detection. For VGG16BN, the proposed method improves TWA in many cases, but FT remains competitive or better for MSP-based detection. For WideResNet-40-2, the gains are also mixed: the proposed method improves Energy-based results on several datasets, while TWA or FT can be stronger for MSP. These results suggest that rank-1 weight-component correction is broadly useful but its effect depends on the backbone and OOD scoring protocol, with the clearest benefit observed for ResNet50 and Energy-based detection.

\resultdatasetwooe

\subsection{Compatibility with Outlier Exposure}
\resulttestaccoe
We first analyze whether the proposed rank-1 correction remains effective when auxiliary outlier data are used, considering both OOD detection and ID classification. Table~\ref{tb:test_acc_oe} reports CIFAR-100 ID accuracy with OE. Although FT achieves the highest ID accuracy, the proposed method improves accuracy over TWA from 69.94\% to 72.69\%, reducing the ID accuracy degradation caused by TWA under OE. Table~\ref{tb:result_w_oe} reports the OOD detection results with OE using ResNet50. Even under explicit outlier supervision, the proposed method consistently improves TWA across all OOD datasets for MSP, Energy Score, and RankWeight. The improvement is particularly clear for Energy-based detection on SVHN, where AUROC increases from 85.4\% to 90.1\% and FPR95 decreases from 75.00\% to 59.89\%. For Energy-based detection on Textures and Places, the gains are more moderate, and FT remains competitive for some FPR95 values. Overall, these results suggest that the proposed weight-space correction is complementary to OE rather than redundant with auxiliary outlier training.

\resultwoe

\subsection{Feature-based OOD Detection}
\label{subsec:feature_based_ood}
We also evaluate feature-based OOD detection. Table~\ref{tb:result_mahalanobis} compares FT, TWA, and the proposed method using Mahalanobis scoring on ResNet50, with features extracted from the input to the final classification layer.

\resultmahalanobis

The results are mixed. On SVHN and Places, the proposed method underperforms TWA in both AUROC and FPR95, with and without OE. For example, without OE on SVHN, AUROC decreases from 84.1\% to 66.3\%, while FPR95 increases from 55.20\% to 86.29\%. On Textures without OE, the proposed method improves AUROC from 86.6\% to 88.2\% and reduces FPR95 from 42.23\% to 41.22\%, outperforming both FT and TWA. With OE on Textures, AUROC improves over TWA by 0.8 percentage points, but FPR95 worsens by 2.40 percentage points; FT achieves the best performance on both metrics.

These results show that the gains observed with MSP and RankWeight do not consistently extend to Mahalanobis-based detection. The effectiveness of rank-1 weight correction therefore depends on the OOD scoring protocol, and improvements in confidence-based detection should not be interpreted as uniform improvements in feature-based detection.

\subsection{Loss Landscape Visualization}
Weight averaging is associated with flatter loss landscapes and improved generalization~\cite{swa,twa}. We examine whether adding rank-1 correction directions to the TWA subspace preserves this flat-solution behavior. Figure~\ref{fig:loss_landscape} visualizes the loss landscapes of FT, TWA, and the proposed method without OE. Compared with FT, both TWA and the proposed method are located in flatter basins, and the proposed method preserves a flat landscape comparable to TWA. This suggests that the proposed rank-1 correction improves OOD detection while retaining the flat-solution property of weight averaging.

\losslandscape

\section{Conclusion}
\label{sec:conclusion}

We proposed rank-1 weight-corrected TWA for OOD detection, augmenting the checkpoint subspace with leading rank-one correction directions learned through the ID classification objective, without changing the architecture or inference cost. On CIFAR-100 with SVHN, Textures, and Places as OOD datasets, the method improves over FT and TWA with MSP and RankWeight on ResNet50, both with and without OE. Mahalanobis results are mixed, demonstrating dependence on the scoring protocol. Analyses favor correction near the classifier and show improvements across checkpoint ranges, with effects varying by backbone. Loss landscape visualization suggests that the method retains the flat-solution behavior of weight averaging.

\paragraph{Discussion and Limitations.}
The proposed method adds SVD-based component extraction and expanded-subspace construction to the TWA optimization stage, but the final model remains a single network and incurs no additional inference-time cost. Our analyses indicate that the effectiveness of the correction depends on the target layer, backbone architecture, and OOD scoring protocol. In particular, the Mahalanobis results show that gains from weight correction do not necessarily translate to feature-space distance-based detection. The current results do not establish the cause of these degradations. Evaluation is limited to CIFAR-100 as ID with SVHN, Textures, and Places as OOD datasets; large-scale and near-OOD evaluation remain outside the scope of this study. In this work, we mainly use the leading rank-one component and select target layers manually; adaptive selection of component ranks and target layers is an important direction for applying the method to larger architectures and more diverse OOD settings.

\bibliographystyle{splncs04}
\bibliography{main}

@String(PAMI  = {IEEE Trans. Pattern Anal. Mach. Intell.})

@String(CVPR  = {IEEE Conf. Comput. Vis. Pattern Recog.})

@String(NeurIPS = {Adv. Neural Inform. Process. Syst.})

@String(ICML  = {Int. Conf. Mach. Learn.})

@String(ICLR  = {Int. Conf. Learn. Represent.})

@String(BMVC  = {Brit. Mach. Vis. Conf.})

@String(PAMI  = {IEEE TPAMI})

@String(CVPR  = {CVPR})

@String(NeurIPS = {NeurIPS})

@String(ICML  = {ICML})

@String(ICLR  = {ICLR})

@String(BMVC  =	{BMVC})

@inproceedings{
msp,
title={A Baseline for Detecting Misclassified and Out-of-Distribution Examples in Neural Networks},
author={Dan Hendrycks and Kevin Gimpel},
booktitle=ICLR,
year={2017},
url={https://openreview.net/forum?id=Hkg4TI9xl}
}

@article{energy,
  title={Energy-based out-of-distribution detection},
  author={Liu, Weitang and Wang, Xiaoyun and Owens, John and Li, Yixuan},
  journal=NeurIPS,
  volume={33},
  pages={21464--21475},
  year={2020}
}

@InProceedings{swa,
  title = 	 {Averaging Weights Leads to Wider Optima and Better Generalization},
  author =       {Izmailov, Pavel and Podoprikhin, Dmitrii and Garipov, Timur and Vetrov, Dmitry and Wilson, Andrew Gordon},
  booktitle = 	 {Uncertainty in Artificial Intelligence},
  pages = 	 {875--884},
  year = 	 {2018},
  volume = 	 {R16},
  series = 	 {Proceedings of Machine Learning Research},
  publisher =    {PMLR},
}

@article{swad,
  title={Swad: Domain generalization by seeking flat minima},
  author={Cha, Junbum and Chun, Sanghyuk and Lee, Kyungjae and Cho, Han-Cheol and Park, Seunghyun and Lee, Yunsung and Park, Sungrae},
  journal=NeurIPS,
  volume={34},
  pages={22405--22418},
  year={2021}
}

@inproceedings{model_soup,
  title={Model soups: averaging weights of multiple fine-tuned models improves accuracy without increasing inference time},
  author={Wortsman, Mitchell and Ilharco, Gabriel and Gadre, Samir Ya and Roelofs, Rebecca and Gontijo-Lopes, Raphael and Morcos, Ari S and Namkoong, Hongseok and Farhadi, Ali and Carmon, Yair and Kornblith, Simon and others},
  booktitle=ICML,
  pages={23965--23998},
  year={2022},
  organization={PMLR}
}

@inproceedings{
    twa,
    title={Trainable Weight Averaging: Efficient Training by Optimizing Historical Solutions},
    author={Tao Li and Zhehao Huang and Qinghua Tao and Yingwen Wu and Xiaolin Huang},
    booktitle=ICLR,
    year={2023},
    url={https://openreview.net/forum?id=8wbnpOJY-f}
}

@article{rankfeat,
  title={RankFeat\&RankWeight: Rank-1 feature/weight removal for out-of-distribution detection},
  author={Song, Yue and Wang, Wei and Sebe, Nicu},
  journal=PAMI,
  volume={47},
  number={4},
  pages={2505--2519},
  year={2024},
  publisher={IEEE}
}

@article{mahalanobis,
  title={A simple unified framework for detecting out-of-distribution samples and adversarial attacks},
  author={Lee, Kimin and Lee, Kibok and Lee, Honglak and Shin, Jinwoo},
  journal=NeurIPS,
  volume={31},
  year={2018}
}

@article{react,
  title={React: Out-of-distribution detection with rectified activations},
  author={Sun, Yiyou and Guo, Chuan and Li, Yixuan},
  journal=NeurIPS,
  volume={34},
  pages={144--157},
  year={2021}
}

@inproceedings{resnet,
  title={Deep residual learning for image recognition},
  author={He, Kaiming and Zhang, Xiangyu and Ren, Shaoqing and Sun, Jian},
  booktitle=CVPR,
  pages={770--778},
  year={2016}
}

@article{ood,
  title={Generalized out-of-distribution detection: A survey},
  author={Yang, Jingkang and Zhou, Kaiyang and Li, Yixuan and Liu, Ziwei},
  journal={International Journal of Computer Vision},
  volume={132},
  number={12},
  pages={5635--5662},
  year={2024},
  publisher={Springer}
}

@article{vos,
      title={VOS: Learning What You Don’t Know by Virtual Outlier Synthesis}, 
      author={Du, Xuefeng and Wang, Zhaoning and Cai, Mu and Li, Yixuan},
      journal=ICLR,
      year={2022}
}

@inproceedings{oe,
  title={Deep Anomaly Detection with Outlier Exposure},
  author={Hendrycks, Dan and Mazeika, Mantas and Dietterich, Thomas},
  booktitle=ICLR,
  year={2019}
}

@misc{vgg16,
    title={Very Deep Convolutional Networks for Large-Scale Image Recognition}, 
    author={Karen Simonyan and Andrew Zisserman},
    year={2015},
    eprint={1409.1556},
    archivePrefix={arXiv},
    primaryClass={cs.CV},
    url={https://arxiv.org/abs/1409.1556}, 
}

@inproceedings{bn,
  title={Batch normalization: Accelerating deep network training by reducing internal covariate shift},
  author={Ioffe, Sergey and Szegedy, Christian},
  booktitle=ICML,
  pages={448--456},
  year={2015},
}

@inproceedings{wideresnet,
  title={Wide Residual Networks},
  author={Zagoruyko, Sergey and Komodakis, Nikos},
  booktitle=BMVC,
  year={2016},
  organization={British Machine Vision Association}
}

@Techreport{cifar100,
    author = {Krizhevsky, Alex and Hinton, Geoffrey},
    address = {Toronto, Ontario},
    institution = {University of Toronto},
    publisher = {Technical report, University of Toronto},
    title = {Learning multiple layers of features from tiny images},
    year = {2009},
    title_with_no_special_chars = {Learning multiple layers of features from tiny images},
    url = {https://www.cs.toronto.edu/~kriz/learning-features-2009-TR.pdf}
}

@misc{tinyimages300k,
    title={Scaling Out-of-Distribution Detection for Real-World Settings}, 
    author={Dan Hendrycks and Steven Basart and Mantas Mazeika and Andy Zou and Joe Kwon and Mohammadreza Mostajabi and Jacob Steinhardt and Dawn Song},
    year={2022},
    eprint={1911.11132},
    archivePrefix={arXiv},
    primaryClass={cs.CV},
    url={https://arxiv.org/abs/1911.11132}, 
}

@inproceedings{svhn,
    title	= {Reading Digits in Natural Images with Unsupervised Feature Learning},
    author	= {Yuval Netzer and Tao Wang and Adam Coates and Alessandro Bissacco and Bo Wu and Andrew Y. Ng},
    year	= {2011},
    URL	= {http://ufldl.stanford.edu/housenumbers/nips2011_housenumbers.pdf},
    booktitle	= {NIPS Workshop on Deep Learning and Unsupervised Feature Learning 2011}}

@inproceedings{dtd,
  title={Describing textures in the wild},
  author={Cimpoi, Mircea and Maji, Subhransu and Kokkinos, Iasonas and Mohamed, Sammy and Vedaldi, Andrea},
  booktitle=CVPR,
  pages={3606--3613},
  year={2014}
}

@ARTICLE{places365,
      author={Zhou, Bolei and Lapedriza, Agata and Khosla, Aditya and Oliva, Aude and Torralba, Antonio},
      journal=PAMI, 
      title={Places: A 10 Million Image Database for Scene Recognition}, 
      year={2018},
      volume={40},
      number={6},
      pages={1452-1464},
      doi={10.1109/TPAMI.2017.2723009}}
\end{document}